\documentclass[letterpaper, 10 pt, conference]{ieeeconf}  

\IEEEoverridecommandlockouts                              

\usepackage{times}
\usepackage{epsfig}
\usepackage{graphicx}
\usepackage{amsmath}
\usepackage{multirow}

\def\eg{\emph{e.g.}} 
\def\ie{\emph{i.e.}} 

\usepackage{amssymb}
\usepackage{pdfpages}

\usepackage{algorithm, algpseudocode}

\definecolor{citecolor}{RGB}{119,185,0} 
\usepackage[pagebackref=false,breaklinks=true,colorlinks,citecolor=citecolor,bookmarks=false]{hyperref}
\usepackage{colortbl}

\newlength\savewidth\newcommand\shline{\noalign{\global\savewidth\arrayrulewidth
  \global\arrayrulewidth 1pt}\hline\noalign{\global\arrayrulewidth\savewidth}}
  
\title{\LARGE \bf
Progressive Text-to-3D Generation for Automatic 3D Prototyping
}
\author{Han Yi$^{1}$, Zhedong Zheng$^{1}$, Xiangyu Xu$^{2}$ and Tat-seng Chua$^{1}$
\thanks{$^{1}$Han Yi, Zhedong Zheng and Tat-seng Chua are with School of Computing, National University of Singapore, Singapore 117417
        {\tt\small hany24@u.nus.edu, \{zdzheng,dcscts\}@nus.edu.sg}}%
\thanks{$^{2}$Xiangyu Xu is with School of Mathematics and Statistics, Xi'an Jiaotong University, China 710049
        {\tt\small xiangyu.xu@xjtu.edu.cn}}%
}

\begin{document}

\maketitle
\thispagestyle{empty}
\pagestyle{empty}


\begin{abstract}
Text-to-3D generation is to craft a 3D object according to a natural language description. This can significantly reduce the workload for manually designing 3D models and provide a more natural way of interaction for users. However, this problem remains challenging in recovering the fine-grained details effectively and optimizing a large-size 3D output efficiently. Inspired by the success of progressive learning, we propose a Multi-Scale Triplane Network (MTN) and a new progressive learning strategy. As the name implies, the Multi-Scale Triplane Network consists of four triplanes transitioning from low to high resolution. The low-resolution triplane could serve as an initial shape for the high-resolution ones, easing the optimization difficulty. 
To further enable the fine-grained details, we also introduce the progressive learning strategy, which explicitly demands the network to shift its focus of attention from simple coarse-grained patterns to difficult fine-grained patterns. Our experiment verifies that the proposed method performs favorably against existing methods. For even the most challenging descriptions, where most existing methods struggle to produce a viable shape, our proposed method consistently delivers. We aspire for our work to pave the way for automatic 3D prototyping via natural language descriptions.

\end{abstract}

\section{Introduction}

3D prototyping, also known as 3D printing or additive manufacturing, is a technology that 
transforms digital models into tangible objects by adding material layer by layer. This technology has revolutionized sectors such as robotics~\cite{hunt20143d, maccurdy2016printable, carrico20173d}, and manufacturing~\cite{drotman20173d, fujii2023hummingbird}. 
However, a significant challenge is that digital design remains time-consuming and labor-intensive. As a solution, researchers are exploring simpler and more intuitive methods to guide 3D generation, like using text prompts. 
The aim of the text-to-3D generation task is to automatically create a 3D object draft from a natural description, thus cutting down the design efforts from the ground up. 

In recent years, text-to-3D generation has reported rapid development due to the breakthrough of 2D text-to-image diffusion models~\cite{dhariwal2021diffusion, nichol2021improved, song2021denoising}. 
For instance, the pioneer work DreamFusion~\cite{poole2022dreamfusion} leverages the 2D Stable Diffusion and proposes Score Distillation Sampling (SDS) algorithm to generate a variety of 3D objects using only text prompts. 
However, there remain two problems. 1) The optimization difficulty of 3D high-resolution objects. 
It is hard to directly map one sentence to one high-dimension 3D object, especially in the form of Neural Radiance Fields (NeRF)~\cite{mildenhall2021nerf}. This leads to either generation collapse or extended training duration for model convergence.
2) 
Lack of fine-grained details.
We notice that some works report blurred results~\cite{poole2022dreamfusion, wang2023score, metzer2023latent}. This is due to the use of a fixed training strategy, i.e., focusing on global fidelity all the time while ignoring the local parts.

\begin{figure}[!t]
\begin{center}
\includegraphics[width=.95\linewidth]{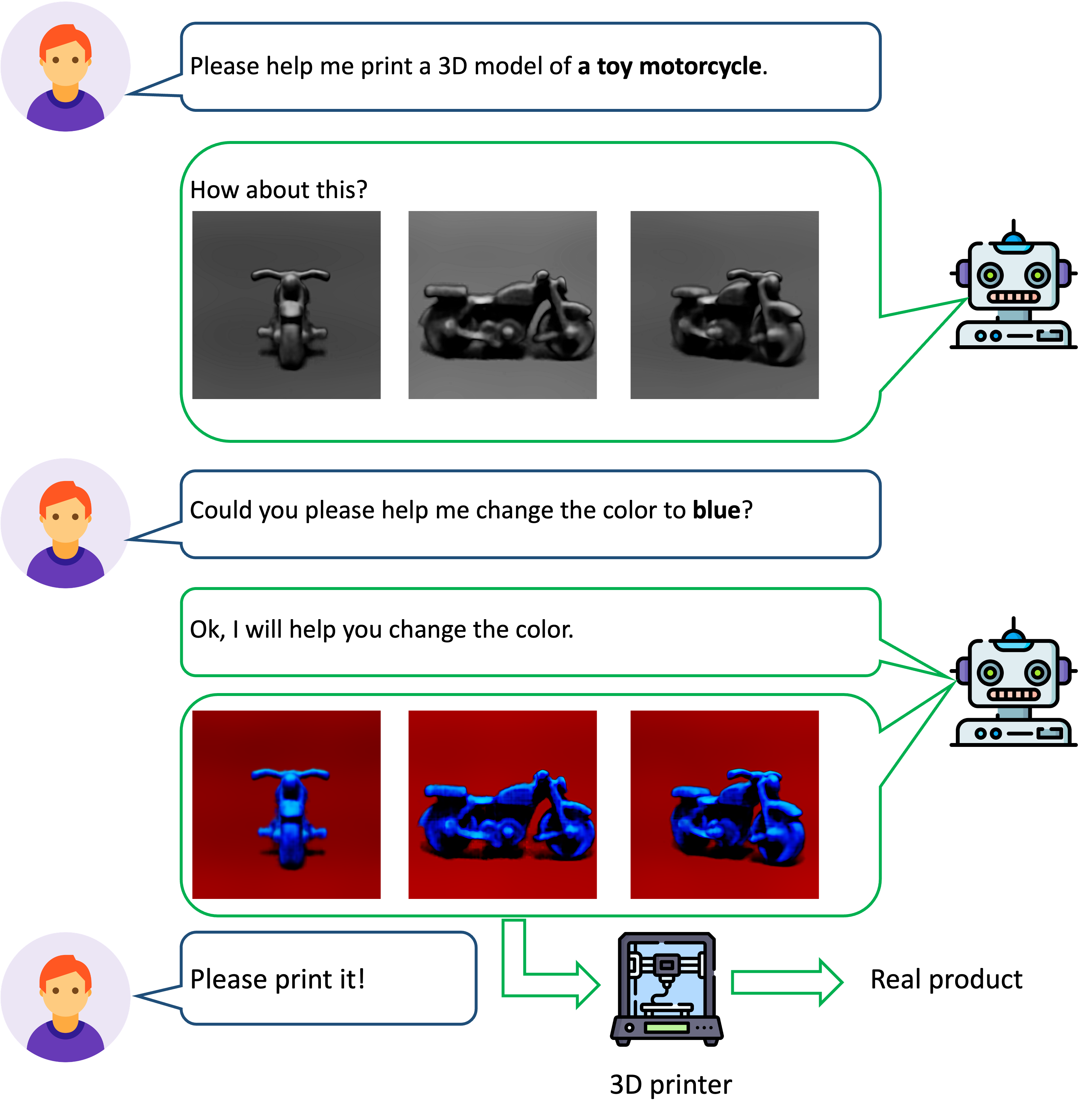}
\end{center}
\vspace{-.1in}
   \caption{Pipeline for fast 3D prototyping. 
   The proposed algorithm facilitates effortless and interactive creation of high-quality 3D objects from natural language descriptions, which can then be utilized for 3D printing. 
   }
\label{fig:first image}
\vspace{-3mm}
\end{figure}

In an attempt to overcome the above-mentioned challenges, we propose a progressive text-to-3D generation model that gradually refines the details of 3D objects. For the first problem, 
we introduce a novel network structure, namely, Multi-Scale Triplane Network (MTN) consisting of four triplanes ranging from low to high resolution. 
In the initial phases of training, we sample low-resolution features from the corresponding low-resolution triplane to capture the basic global geometric shape. As training advances, we fix the former low-resolution triplanes and gradually shift our focus to triplanes with a higher resolution.
Such a progressive structure facilitates the model to capture different-level features in a step-by-step manner and thus enhances the geometric and textural nuances of the 3D model, such as color and texture.

For the second problem, we adopt a progressive learning strategy focusing on two key factors, \ie, time step $t$ and camera radius. 
In particular, unlike existing 2D diffusion models that utilize random sampling, we adopt a large $t$ during the initial stages to guide the global structure. As the training progresses, we transition to a smaller 
$t$ to refine visual details. 
Meanwhile, we gradually adjust the radius of the camera to approach the object more closely. This enables the camera to initially focus on capturing the global structure and later shift its attention to the local details.

To summarize, our contributions are as follows: 
\begin{itemize}
    \item 
    Most existing works on text-to-3D struggle to craft high-resolution outputs due to the optimization difficulty. In contrast, we introduce a Multi-Scale Triplane Network (MTN) to gradually create the 3D model in a bottom-up style, effectively alleviating the optimization issue.
    %
    \item We also propose a progressive learning strategy that simultaneously reduces the camera radius and
    time step $t$ in diffusion to refine details of the 3D model in a coarse-to-fine manner. 
    \item Albeit simple, extensive experiments show that the proposed method could achieve high-resolution outputs that align closely with natural language descriptions. 
    We expect this work to pave the way for automatic 3D prototyping via easier human-machine interaction.
\end{itemize}

\section{Related Work}
\subsection{Text-Guided Diffusion Models}

The evolution of text-to-image generative paradigms has been notably characterized by the advent of diffusion models~\cite{dhariwal2021diffusion, nichol2021improved, song2021denoising}. These models have gained prominence for their robust stability and unprecedented scalability. Significant exemplars in this category include GLIDE~\cite{nichol2021glide}, DALL·E 2~\cite{ramesh2022hierarchical}, Imagen~\cite{saharia2022photorealistic}, and Stable Diffusion~\cite{rombach2022high}. These models leverage textual prompts to synthesize images of photorealistic quality. Their efficacy is substantially bolstered by the availability of comprehensive datasets comprising billions of image-text pairs, thereby enhancing the semantic understanding of these generative systems. 
However, it is nontrivial to extend these diffusion models to 3D generation, which needs to accurately synthesize 3D attributes from textual prompts across a diversity of viewpoints~\cite{zhang2023multi}.

\subsection{3D Generative Modeling} 
The realm of 3D generative modeling has seen extensive exploration across diverse representation types, including voxel grids~\cite{tatarchenko2017octree, li2017grass}, point clouds~\cite{luo2021surfgen, zhou20213d, vahdat2022lion}, meshes~\cite{gao2019sdm, gao2021tm, nash2020polygen, henderson2020leveraging, gupta2020neural,rosinol2019incremental}, implicit fields~\cite{cheng2022cross, wu2020pq, wu2022learning, zheng2022sdf}, and octrees~\cite{ibing2023octree}. While many traditional approaches hinge on 3D assets as training data, the challenge of acquiring such data at scale has spurred alternative strategies. 
Addressing the inherent challenge of obtaining 3D assets for training, some recent endeavors have turned to 2D supervision. Leveraging ubiquitous 2D images, models like pi-GAN~\cite{chan2021pi}, EG3D~\cite{chan2022efficient}, MagicMirror~\cite{zheng2022magic} and GIRAFFE~\cite{niemeyer2021giraffe} have supervised 2D renderings of 3D models through adversarial loss against 2D image datasets. While these approaches hold potential, a recurring challenge is that they are often restricted to specific domains like human faces~\cite{karras2019style}, 
limiting their versatility and hindering expansive creative freedom in 3D design. In our study, we pivot towards text-to-3D generation, with the objective of crafting a visually favorable 3D object guided by diverse text prompts.

\subsection{Text-to-3D Generation}

The success of the text-to-image generation models has driven substantial progress in the emerging field of text-to-3D object generation.
Notably, the integration of CLIP into models like Dream Fields~\cite{jain2022zero} and CLIPmesh~\cite{mohammad2022clip} has been a significant advancement.
These approaches harness CLIP to optimize 3D representations, ensuring 2D renderings resonate with textual prompts. A defining advantage of such techniques is their ability to bypass the need for costly 3D training data, though a trade-off in terms of the realism of the resultant 3D models has been observed.
More recent advancements, such as DreamFusion~\cite{poole2022dreamfusion}, SJC~\cite{wang2023score}, Magic3D~\cite{lin2023magic3d}, and Latent-NeRF~\cite{metzer2023latent}, have showcased the merits of employing robust text-to-image diffusion models as a robust 2D prior, elevating the quality and realism of text-to-3D generation. This innovation, capitalizing on the potential of diffusion models, has led to outcomes with higher fidelity and diversity, as well as reduced generation time. We 
follow the spirit of this line of work and 
present new techniques to effectively improve the quality of the 3D outputs.

\section{Method}

\begin{figure*}[t]
\begin{center}
\includegraphics[width=1.0\linewidth]{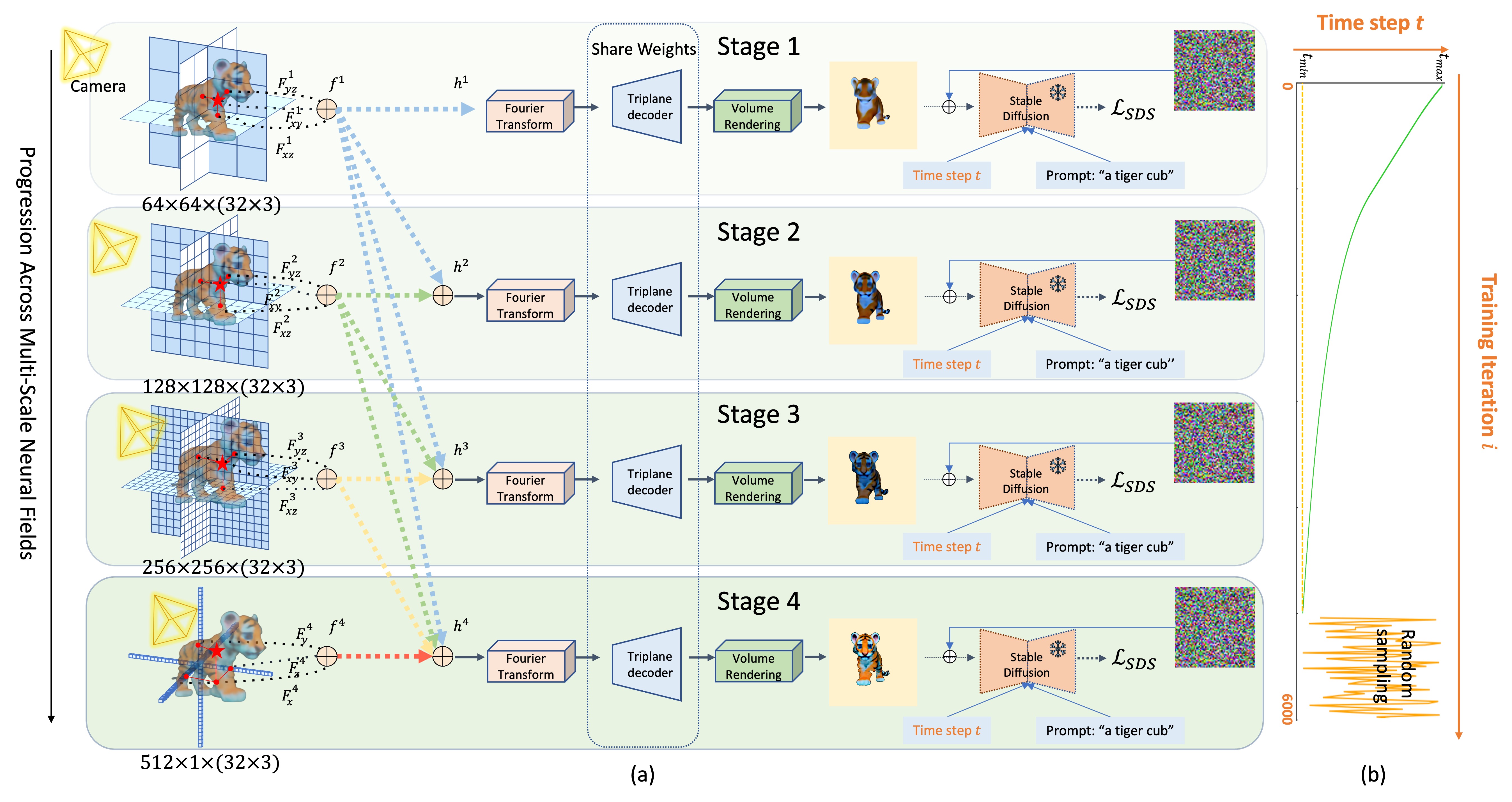}
\end{center}
\vspace{-.2in}
   \caption{Overview of the proposed Multi-Scale Triplane Network (MTN). \textbf{(a)} Given the text prompts like ``a tiger cub'', MTN produces 3D representations using Multi-Scale Neural Fields, utilizing four triplanes varying in resolution. To save memory costs and enable the highest resolution, we make a trade-off to deploy the high-dimension trivector format as the triplane alternative. First, by casting rays from a random camera position and view, we can sample a lot of 3D points along each ray and then encode their corresponding features by projecting them onto triplanes. 
   After the 3D input encoding, the network uses a Fourier transform, a triplane decoder, and volume rendering. The Fourier feature transform~\cite{tancik2020fourier} enables the triplane decoder to learn high-frequency information. The triplane decoder is a shallow MLP. Volume rendering is a technique~\cite{mildenhall2021nerf} that converts the 3D representation into RGB images. 
   In particular, our training process unfolds in four progressive stages. Initially, we leverage the low-resolution triplanes, tapping into their global geometric insights. As training advances, we fix the former low-resolution triplanes, while updating and drawing from the higher-resolution triplanes. This shift refines intricate details in the 3D model, capturing subtle textures and shading. \textbf{(b)} Concurrently, as training proceeds, the time step $t$ undergoes progressive adjustments, and the camera also approaches the neural field progressively, emphasizing the refinement of local features. To update parameters, we employ a frozen Stable Diffusion model to estimate the injected noise on the rendered image (\eg, tiger) and then backpropagate the gradient. }
\label{fig:pipeline}
\vspace{-3mm}
\end{figure*}

In this section, we first explain our Multi-Scale Triplane Network (MTN) with discussion. Then we elaborate a progressive learning strategy, followed by implementation details. The brief scheme is shown in Fig.~\ref{fig:pipeline}.  

\subsection{Multi-Scale Triplane}

An overview of our Multi-Scale Triplane Network (MTN) is shown in Fig.~\ref{fig:pipeline}.
In particular, MTN 
is composed of four triplanes~\cite{chan2022efficient} ranging from low to high resolutions. 
Each triplane leverages three axis-aligned 2D feature planes $\mathbf{F}^{m}_{x y}, \mathbf{F}^{m}_{x z}, \mathbf{F}^{m}_{y z} \in \mathbb{R}^{N_m \times N_m \times C}, m=1,2,3$. $N_m$ denotes spatial resolution, while $C$ is the dimension of channels and $m$ represents the training stage. 
Note that a large $N_m$ results in a substantial GPU memory cost. Therefore, for the last triplane, we essentially employ a trivector instead to optimize memory usage and support higher resolution.
%
This trivector configuration leverages the axis-aligned vectors $\mathbf{F}^4_{x}, \mathbf{F}^4_{y}, \mathbf{F}^4_{z} \in \mathbb{R}^{N_4 \times 1 \times C}$ with a resolution of $N_4 \times 1$ and $C$. 

Given any 3D coordinate point $p \in \mathbb{R}^{3}$, we project this coordinate onto each of these orthogonal feature 
planes and sample feature vectors via interpolation. We then sum these three vectors $f^{m}(p) = \mathbf{F}^{m}_{x y}(p)+\mathbf{F}^{m}_{x z}(p)+ \mathbf{F}^{m}_{y z}(p)$ for $m=1,2,3$ as position features for the first three triplanes, while $f^{4}(p) = \mathbf{F}^{4}_{x}(p)+\mathbf{F}^{4}_{y}(p)+ \mathbf{F}^{4}_{z}(p)$ for the last trivector.
To aggregate multi-scale features, we further fuse the different level position features together as $h^{m}(p) = \sum_{k=1}^{m}(f^{k}(p))$. 
After obtaining the multi-scale representation, we follow \cite{tancik2020fourier} to transform the summed position features into the Fourier domain.
Subsequently, the Fourier features are fed forward into a lightweight triplane decoder to estimate color and density~\cite{mildenhall2021nerf}. 
We deploy a Multi-Layer Perceptron (MLP) as the triplane decoder. 
Finally, to calculate the loss, we apply neural volume rendering techniques~\cite{mildenhall2021nerf} to project the 3D representation onto an RGB image $I$, which is the input of the Diffusion model.

\noindent\textbf{Discussion. Why is multi-scale important?} As shown in Fig.~\ref{fig:pipeline}, we utilize triplanes with different resolutions to obtain features at different scales. We intend to simulate the human recognition system 
to transition from basic elements to more intricate details when observing 3D objects.
For instance, when a human sees a new object, he will first capture the overarching structure of the model and then refine the details via the foveal vision. 
Similarly, during the early stages of training, we extract low-resolution features from the corresponding low-resolution triplane.
As one point on a low-resolution triplane is obtained through interpolation from a coarse grid and encompasses a broader field of view, these coarse features provide inherent global geometric insights. 

As the training progresses, we gradually shift our focus from the initial low-resolution triplanes to the higher-resolution counterparts. The high-resolution triplane could acquire intricate features to refine details in the 3D model, such as subtle shade and texture nuances. 
Simultaneously, it is easier to refine high-scale features and facilitate the optimization process of high-scale features if low-scale features have been already well optimized. 
This process is also in spirit similar to the classical curriculum learning~\cite{bengio2009curriculum} where learning begins with simpler tasks and gradually advances to harder ones.
In the experimental evaluations, we could observe that the proposed method achieves visual enhancements in both the shape and texture of the model, even for complex descriptions.

\noindent\textbf{Optimization objective.} 
Given the projected image $I$, we apply Score Distillation Sampling (SDS)~\cite{poole2022dreamfusion} to distill 2D image priors from the pretrained 2D diffusion model $\boldsymbol{\epsilon}_\phi$. The loss on 2D projection is then back-propagated to update differentiable 3D representations. 
In particular, the proposed 3D model can be typically depicted as a parametric function $I = g_{\theta}(P)$, where \( I \) represents the images produced at distinct camera poses, and $P$ is the set of multiple positions $p$. Here, \( g \) denotes the volumetric rendering mechanism, and \( \theta \) embodies a coordinate-based MLP and triplanes that portray a 3D scene. To estimate the projection quality, we adopt the pretrained diffusion model, which is well aligned with text prompts $y$. The one-time denoising forward can be formulated as  
\( \epsilon_\phi(I_t; y, t) \) to predict the noise \( \varepsilon \) given the noisy image \( I_t \), time step $t$, and text embedding \( y \). 
Therefore, the gradient of the SDS loss can be formulated as:
$$\nabla_\theta \mathcal{L}_{SDS}(\phi, g_{\theta}(P))=\mathbb{E}_{t, \epsilon}\left[\left(\epsilon_\phi\left(I_t ; y, t\right)-\epsilon\right) \frac{\partial I_t}{\partial \theta}\right], $$
where $\boldsymbol{\epsilon}$   is a noise term following a standard normal distribution 
and $I_t$ denotes the noisy image. Following the setting in the diffusion model~\cite{dhariwal2021diffusion, nichol2021improved, song2021denoising}, the noisy image can be formulated as a linear process $I_t=\sqrt{\bar{\alpha}_t} I+\sqrt{1-\bar{\alpha}_t} \boldsymbol{\epsilon}$, where 
$\bar{\alpha}_t$
is a predefined time-dependent constant. Besides, it is worth noting that the diffusion model parameter \( \phi \) is frozen. The purpose of this denoising function is to offer the 
text-aware guidance to update \( \theta \). If the projection $I$ is well-aligned with the text $y$, the noise on $I_t$ is easy to predict. Otherwise, we will punish the 3D model. 

\subsection{Progressive Learning Strategy}
Another essential element underlying the proposed method is the employment of a progressive learning strategy, focusing on two critical parameters: \ie, the time step $t$ and camera radius. Subsequent sections provide detailed illustrations for each of these components.

\noindent\textbf{Progressive time step sampling.} 
We first introduce a progressive time step $(t)$ sampling approach. It is motivated by the observation that the default uniform $t$-sampling in SDS training often results in inefficiencies and inaccuracies due to the broad-range random sampling. 
Our approach, therefore, emphasizes a gradual reduction of the time step, directing the model to transition from coarse to detailed learning (See Fig.~\ref{fig:pipeline} (b)).
In the early phases of training, we adopt larger time steps to add a substantial amount of noise into the image. During the noise recovery process, the network is driven to focus on the low-frequency global structure signal.
As training advances and the global structure stabilizes, we decrease to smaller time steps. In this stage, the network is demanded to recover the high-frequency fine-grained pattern according to the context. It facilitates the model in refining local details, such as textures and shades.

We define the rate of change of variable $t$ as follows:
\begin{align}\label{eq:t_derivative}
\frac{\mathrm{d}t}{\mathrm{d}i} = \beta v(t),
\end{align}
where $v(t)$ controls how $t$ changes with respect to the training iteration $i$ and is manually designed. $\beta$ is a positive constant.
We define $v(t)$ piece-wise:
\begin{align}
    v(t) &= \begin{cases} -\exp(\frac{t-n_2}{m_2} ) & \text { if } t>n_2 \\ -1.0 & \text { if } n_1 \leq t \leq n_2 \\ -\exp(\frac{t-n_1}{m_1} ) & \text { if } t<n_1,\end{cases}.
\end{align}
Here, $v(t)<0$ implies $\frac{\mathrm{d}t}{\mathrm{d}i}<0$, indicating that $t$ decreases as training progresses.
Our design ensures that $t$ decreases rapidly at the beginning ($t>n_2$), linearly in the middle ($n_1 \leq t \leq n_2$), and more mildly towards the end ($t<n_1$).
%
%
%
%
After the time step $t$ decreases to $t_\text{min}$, we revert to random sampling:
$$
t \sim \mathcal{U}(t_{\text{min}},t_{\text{max}}). \nonumber
$$
It reintroduces randomness to maintain the vibrancy of the coloration of the 3D model.
%
%
%

We notice that a concurrent pre-print work, Dreamtime~\cite{huang2023dreamtime}, also employs a similar non-increasing $t$-sampling strategy. 
However, such a strategy sometimes tends to overfit the local details, and inadvertently change the global illumination.
Therefore, it is crucial to avoid the consistent use of extremely small time steps at the end of training.
Unlike Dreamtime~\cite{huang2023dreamtime}, our method decreases $t$ with the training step at a much steeper pace and employs a mixture of both deterministic and random sampling.

\noindent\textbf{Progressive radius.}
Simultaneously, our approach also incorporates a dynamic camera radius considering the camera movements in the real world. Typically, eyes will move closer for detailed object observation. Motivated by this behavior, we dynamically adjust the camera radius during the multi-scale learning.
During the low-scale triplane stage, which focuses on broader geometric structures, we utilize a large camera radius to cover the entire object. 
As we move to the high-scale triplane stage, which refines local model details, the camera radius is reduced to closely focus on finer details of the 3D scene. 
This progressive radius strategy is intuitive and directly impacts resolution, aiding in feature learning across varying scales. In the ablation study, we also verify the effectiveness of this strategy (See Section~\ref{sec:ablation}). 

\begin{figure*}[!t]
\begin{center}
\includegraphics[width=0.95\linewidth]{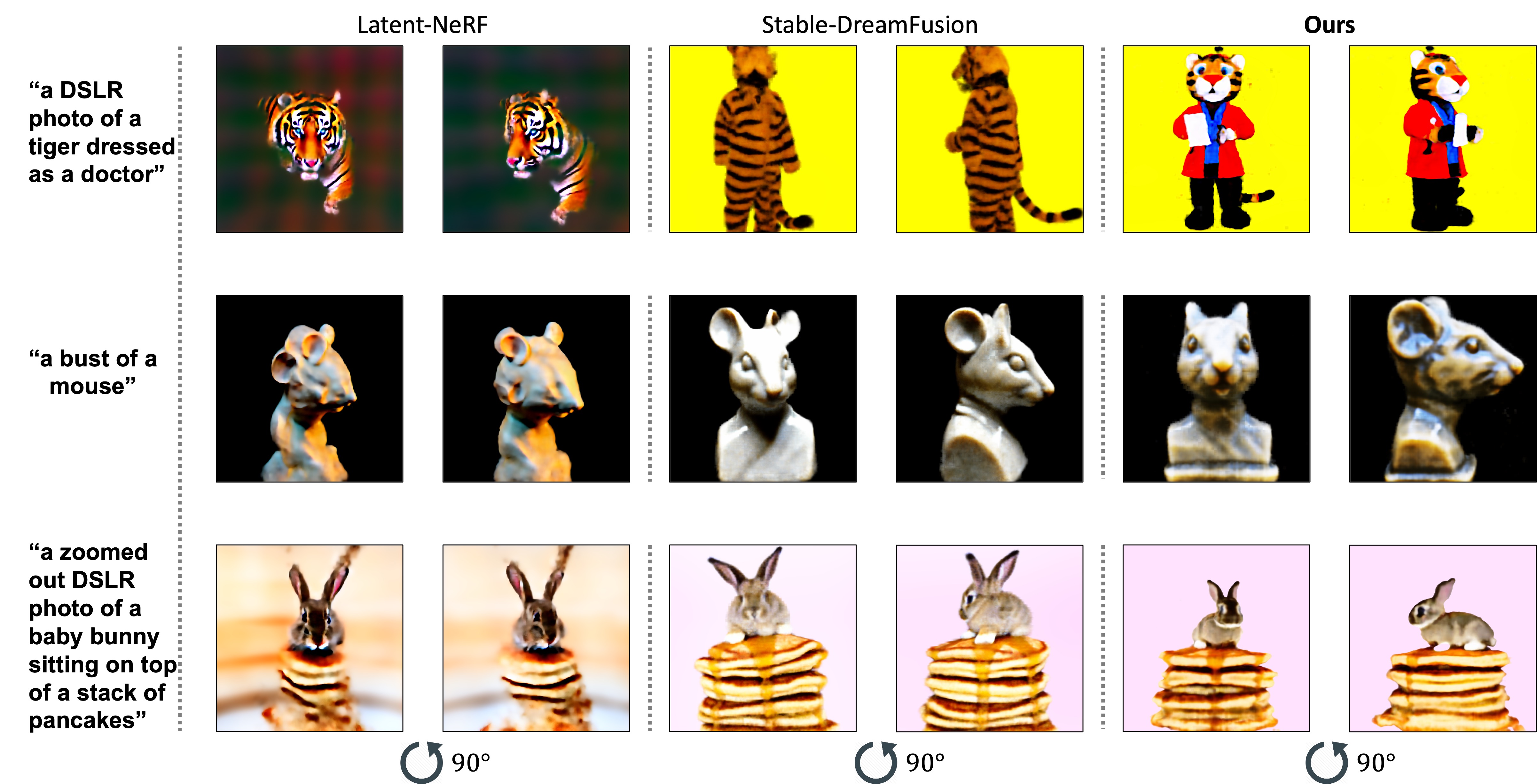}
\end{center}
\vspace{-.1in}
   \caption{Qualitative comparisons for text-to-3D generation among our method, Latent-NeRF~\cite{metzer2023latent}, and Stable-DreamFusion~\cite{stable-dreamfusion}. Here we show the 2D projection of the front view and side view of the 3D model. All results are produced using a fixed random seed. We can observe that the proposed method could generate a higher-fidelity 3D representation aligned with the given description, reducing the extra post-processing costs.}
\label{fig:qualitative comparison}
\end{figure*}

\subsection{Implementation Details}

\noindent\textbf{Neural field rendering structure.} 
As mentioned, the proposed MTN consists of three triplanes and one trivector varying in resolution. 
The resolutions of the triplanes are $N_1,N_2,N_3 = 64, 128, 256$, and the number of channels is $C = 32$. 
For the trivector, we set $N_4 = 512$. 
During the Neural Field optimization, camera positions are randomly sampled in spherical coordinates. 
The azimuth angles, the polar angles and fovy range are randomly sampled between $\left[-180^{\circ}, 180^{\circ}\right]$,  $\left[45^{\circ}, 105^{\circ}\right]$, and $\left[10^{\circ}, 30^{\circ}\right]$, respectively.
For spherical radius, the initial
\( R \in [3.0, 3.5] \) and gradually 
decreases to \( R \in [1.8, 2.1] \).

\noindent\textbf{Diffusion model.}
We deploy DeepFloyd IF~\cite{DeepFloyd-IF} as the guidance model to provide 2D image priors. 
For time step $(t)$ sampling in SDS, the Stable-DreamFusion uses random sampling $t \sim \mathcal{U}(20,980)$. In our proposed approach, the time step $t$ is set to decrease from 980 to 20. Through a grid search, we empirically set an optimal prior weight configuration as \( \{m_1=50, m_2=150, n_1=500, n_2=800\} \) to control the rate of decrease.
Following existing works~\cite{poole2022dreamfusion, metzer2023latent, lin2023magic3d, armandpour2023re}, we also adopt the viewpoint-aware prompts by appending prompts such as ``front view'', ``side view'', and ``back view''.

\noindent\textbf{Optimization.} The number of total iterations is 6000 and the batch size is 1. We employ the Adan optimizer~\cite{xie2022adan} with learning rate of \(1 \times 10^{-3}\), 
weight decay of \(2 \times 10^{-5}\). 
We follow existing works~\cite{chan2022efficient} and apply two regularization terms, \ie, TV regularization and L2 regularization, to prevent floating clouds.

\section{Experiment}
In this section, we assess the capability of our method to produce high-fidelity 3D objects 
according to natural language prompts. 
We primarily consider two key evaluation aspects: (1) alignment with the text, particularly focusing on key words in the sentence; and (2) consistent geometric shape, especially in localized parts like ears and tails.
Due to the space limitation, we mainly compare our approach against two widely-used text-to-3D frameworks. Since  DreamFusion~\cite{poole2022dreamfusion} is not publicly available, we utilize the open-source variant, Stable-DreamFusion~\cite{stable-dreamfusion}. 
Besides, we also compare the proposed method with the pioneer work, \ie, Latent-NeRF~\cite{metzer2023latent}. 

\begin{table}[t]
\scriptsize
\vspace{-.1in}
\begin{center}
{
\setlength{\tabcolsep}{5pt}
\caption{Quantitative comparisons with ground-truth images, Latent-NeRF~\cite{metzer2023latent}, Stable-Dreamfusion~\cite{stable-dreamfusion} evaluated on 153 standard prompts in Dreamfields~\cite{jain2022zero}. The best precision in every column is in \textbf{bold}.}
\label{table:quantative comparison}
\begin{tabular}{l|cccccc}
\shline
\multirow{3}{*}
{Method} & \multicolumn{6}{c}{ R-Precision (\%) $\uparrow$} \\ 
& \multicolumn{2}{c}{CLIP B/32} & \multicolumn{2}{c}{CLIP B/16} & \multicolumn{2}{c}{CLIP L/14} \\ 
& RGB & DEPTH & RGB & DEPTH & RGB & DEPTH \\
\hline
GT images &77.1 & - & 79.1 & - & - &  -\\
Latent-NeRF &48.4 &  37.1 &  52.9 & 40.6 & 59.5 & 40.9  \\
Stable-Dreamfusion &56.4 &  45.9 &  60.3 & 45.8 & 58.3 & 42.9 \\
\textbf{Ours} &\textbf{62.6} & \textbf{53.1} & \textbf{62.6} & \textbf{51.9}  & \textbf{64.8} & \textbf{47.6}  \\
\shline
\end{tabular}
}
\end{center}
\vspace{-3mm}
\end{table}

\begin{table*}[!t]
\scriptsize
\centering
\setlength{\tabcolsep}{9pt}
\caption{Ablation study of different components. The best precision in every column is in \textbf{bold}.
} \label{table:ablation study}
\vspace{-.1in}
\begin{tabular}{l|c|c|c|c|c|c|c|c|c|c|c}
\shline
\multirow{3}{*}{Method} & \multirow{3}{*}{MTN} & \multirow{3}{*}{Progressive} & \multirow{3}{*}{Progressive} & \multicolumn{8}{c}{R-Precision (\%) $\uparrow$} \\
& &  &  & \multicolumn{2}{c}{CLIP B/32} & \multicolumn{2}{c}{CLIP B/16} & \multicolumn{2}{c}{CLIP L/14} & \multicolumn{2}{c}{Mean} \\
& & time step & radius & RGB & DEPTH & RGB & DEPTH & RGB & DEPTH & RGB & DEPTH \\
\hline
Single triplane &  &  & & 46.8	&38.4	&51.8	&41.1	&53.9	&41.4 & 50.8 & 40.3\\
MTN &$\checkmark$&   &   & 57.8&	46.7&	58.2	&46.2	&62.2	&42.8 & 59.4 & 45.2\\
MTN-t &$\checkmark$ &  $\checkmark$ &   & 60.2 & 52.7 & 61.2 & 51.0 & 63.5 & 43.5 & 61.6 & 49.1\\
MTN-r &$\checkmark$ &   & $\checkmark$ &  57.9 &48.5 & 60.4&	48.8	&62.4	&42.7 & 60.2 & 46.7 \\
\textbf{MTN-full} &$\checkmark$ &  $\checkmark$ & $\checkmark$ & \textbf{62.6}&	\textbf{53.1}&	\textbf{62.6}	&\textbf{51.9}&	\textbf{64.8}&	\textbf{47.6} & \textbf{63.3} & \textbf{50.9}\\
\shline
\end{tabular}
\vspace{-3mm}
\end{table*}


\subsection{Qualitative Evaluation}

As shown in Fig.~\ref{fig:qualitative comparison}, we could observe that our method 
outperforms prior competitive approaches in terms of text alignment, texture details, and geometric precision. 
For instance, in the first row, Latent-NeRF~\cite{metzer2023latent} struggles to generate a 3D model. Stable-DreamFusion~\cite{stable-dreamfusion} does generate a tiger avatar but misses the key word ``doctor''. In contrast, the proposed method successfully crafts a tiger doctor with a book in his hands.
In the second row, our method displays a better geometric shape and correct shading on the bust, while Stable-DreamFusion places a tail on the head and Latent-NeRF produces a head shape with three ears.
In the third row, our method successfully captures the keyword ``baby bunny'' and shows a better geometric shape with one head and two ears.
In contrast, 
Latent-NeRF~\cite{metzer2023latent} and Stable-DreamFusion~\cite{stable-dreamfusion} are both plagued with the multi-face and multi-ear issue. 
In summary, our method can generate reliable 3D representations that are aligned with the text prompts and exhibit natural geometric shapes, which are also well-aligned with human intuition. 

\subsection{Quantitative Evaluation}
Since our task is a generation problem, we do not have 3D ground-truth meshes for direct comparison of differences. Therefore, we follow the existing work, \ie, DreamFusion~\cite{poole2022dreamfusion} to evaluate the alignment between 2D projected images and the text prompt. 
In particular, we adopt the CLIP R-Precision~\cite{radford2021learning} to evaluate the retrieval performance for both RGB images and depth maps. The RGB images serve as an indicator of texture quality, while the depth maps represent the geometric shape. A higher score indicates better performance. This evaluation is conducted using three pre-trained CLIP models with different model sizes \ie, CLIP B/32, CLIP B/16, and CLIP L/14.
For a fair comparison, we also adopt 153 standard prompts from Dreamfields~\cite{jain2022zero}.  Our results are shown in Table~\ref{table:quantative comparison}. We could see that our method consistently achieves the highest R-Precision scores across all three metrics, indicating a significant advantage.

\subsection{User Study}
\begin{figure}[t]
\begin{center}
\includegraphics[width=0.95\linewidth]{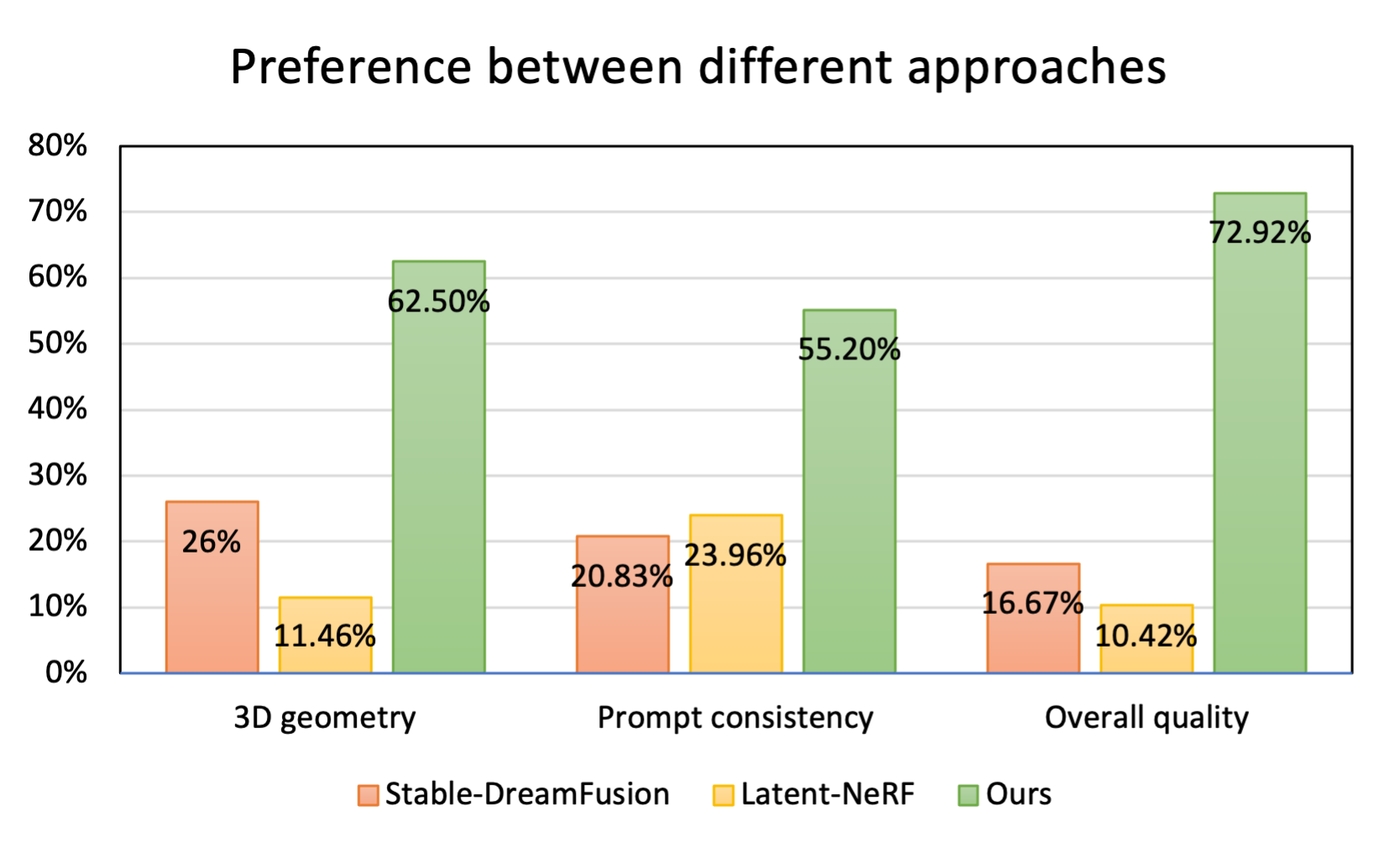}
\end{center}
\vspace{-.1in}
   \caption{User study on visual quality. We could observe that 
   the proposed method excels in 3D geometry and closely aligns with user prompts, and outperforms the two competitive approaches in overall quality.
   }
\vspace{-3mm}
\label{fig:user study}
\end{figure}
For a more comprehensive evaluation, we conduct a user study with 96 participants. We evaluate our model against two prevailing approaches 
like Latent-NeRF~\cite{metzer2023latent} and Stable-DreamFusion~\cite{stable-dreamfusion} in three key aspects: 3D geometry, prompt consistency, and overall quality. We randomly select 96 prompts from the standard set of 153 prompts and generate 3D models using Stable-DreamFusion~\cite{stable-dreamfusion}, Latent-NeRF~\cite{metzer2023latent}, and our approach. Participants are then asked to rank the models based on the aforementioned criteria. 
As shown in Fig.~\ref{fig:user study}, our visual results outperform other methods across multiple metrics, attracting preferences from $72.92\%$ of participants for overall quality, $62.5\%$ for 3D geometry, and $55.2\%$ for prompt consistency. This highlights the efficacy of our approach in delivering superior results across various evaluation criteria.

\subsection{Ablation Study} \label{sec:ablation}
\begin{figure}[t]
\begin{center}
\includegraphics[width=1.0\linewidth]{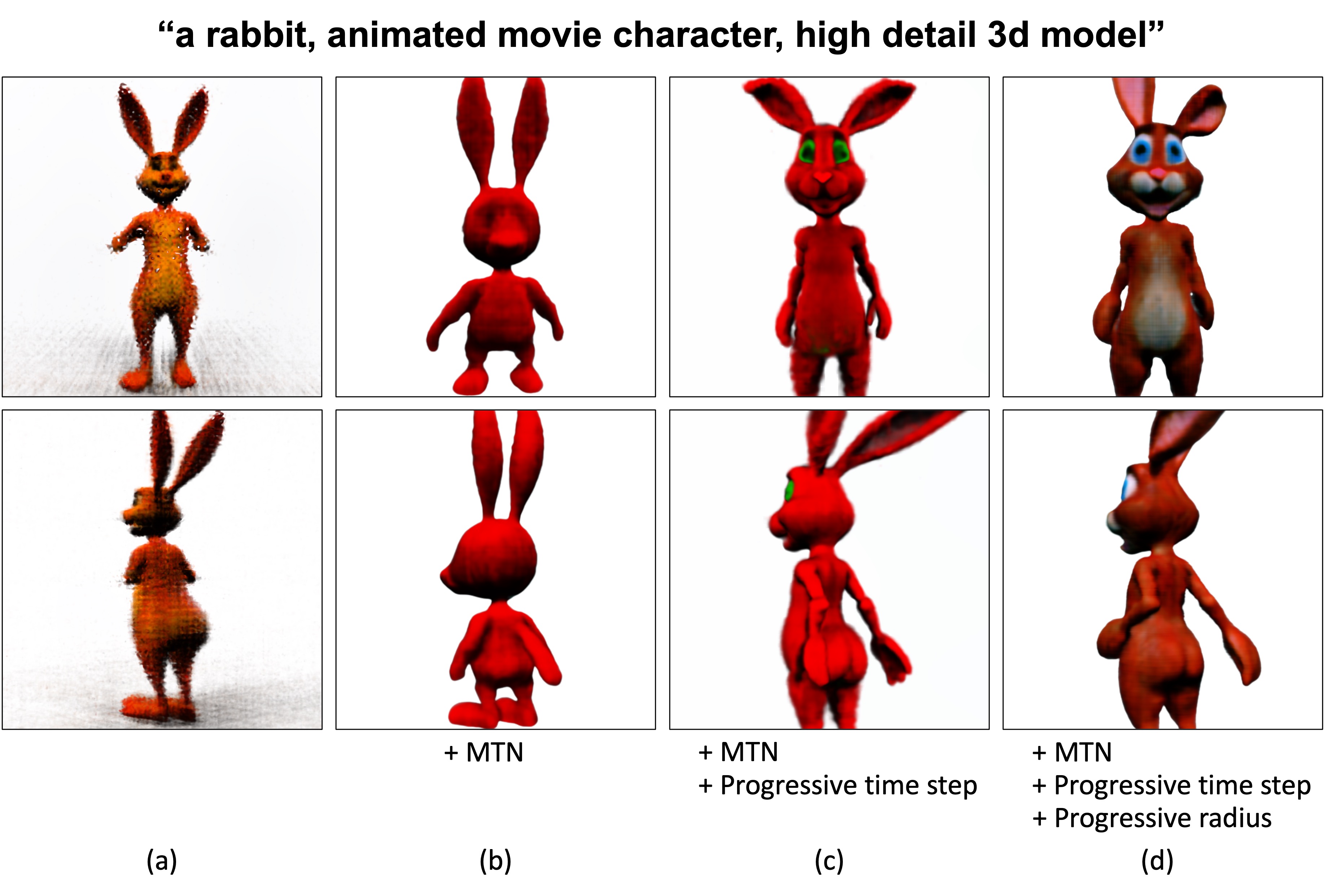}
\end{center}
\vspace{-.1in}
   \caption{Impact of progressive learning strategy. (a) Single triplane; (b) add MTN; (c) add Progressive time step strategy; (d) add Progressive radius, our full method.}
\label{fig:ablation comparison}
\vspace{-.2in}
\end{figure}

\noindent\textbf{Effectiveness of multi-scale triplanes.} 
We first investigate the impact of the multi-scale triplane architecture to substantiate its advantages. As shown in Table~\ref{table:ablation study}, we could observe that the multi-scale architecture facilitates both texture and geometric shape learning. Specifically, the RGB R-Precision is improved with a large margin  $+8.6\%$ on average, while the mean depth R-Precision increases $+4.9\%$. We also provide a visualization result in Fig.~\ref{fig:ablation comparison} (b). The basic single-scale triplane structure results in a 3D output that misses intricate details both texturally and geometrically, evident in incomplete hands, tails, and the presence of floating points. In contrast, the multi-scale network gradually leverages the multi-scale information, yielding a more smooth geometric shape with clear edges. 

\noindent\textbf{Effectiveness of progressive learning.}  Here we further evaluate the impact of progressive time step sampling and progressive radius.  
(1) As shown in the third row of Table~\ref{table:ablation study}, the MTN with only progressive time step strategy could further improve the text alignment by $+2.2\%$ texture and $+3.9\%$ geometry quality on average. This is because the small time step towards the end of learning shifts the focus to high-frequency details, significantly improving the overall visual quality.
As shown in the third column in Fig.~\ref{fig:ablation comparison}, we could notice more fine-grained texture patterns are generated on the surface.  
(2) Similarly to how humans often take a closer look to examine object details, our model, when applying the progressive radius approach, performs even better, showing a $+1.7\%$ improvement on the local texture details. As the camera gets closer, the 2D projection and the optimization objects both emphasizes local quality, resulting in a refined 3D model. 
As a result, the culmination of these strategies leads to a final output that is both detailed and visually appealing (See Fig.~\ref{fig:ablation comparison} (d)).

\section{Conclusion}
In this work, 
inspired by the bottom-up spirit, we introduce the Multi-Scale Triplane Network (MTN) and a progressive learning strategy, both of which effectively ease the optimization difficulty during high-fidelity generation. 
The Multi-Scale Triplane Network operates at the structure level to aggregate the multi-scale representation, while the progressive learning strategy functions at the recognition level to gradually refine high-frequency details.
Extensive experiments verify the effectiveness of every component.
We envision our approach offers  a 
preliminary attempt for automatic 3D prototyping, bridging the gap between natural language descriptions and intricate 3D design. 
In the future, we will continue to explore the potential to complete occluded 3D objects~\cite{mohammadi20233dsgrasp} via language prior and discriminative language guidance~\cite{matsuzawa2023question}.






\bibliographystyle{IEEEtran}
\bibliography{root}

\end{document}